\documentclass{article}

\usepackage{arxiv}

\usepackage[utf8]{inputenc} 
\usepackage[T1]{fontenc}    
\usepackage{hyperref}       
\usepackage{url}            
\usepackage{booktabs}       
\usepackage{amsfonts}       
\usepackage{nicefrac}       
\usepackage{microtype}      
\usepackage{lipsum}		    
\usepackage{graphicx}
\usepackage{pifont}
\usepackage[sort,numbers]{natbib}
\usepackage{doi}

\usepackage{outlines}
\usepackage{acro}
\usepackage{orcidlink}
\usepackage{booktabs}
\usepackage{subcaption}
\usepackage{siunitx}
\usepackage{multirow}
\usepackage{balance}
\usepackage[super]{nth}
\usepackage{algorithm}
\usepackage{algpseudocode}

\usepackage[figuresright]{rotating}
\usepackage{xcolor}

\hypersetup{
pdftitle={EXCODER: EXplainable Classification Of DiscretE time series Representations},
pdfauthor={Yannik Hahn, Antonin Königsfeld, Hasan Tercan, Tobias Meisen},
pdfkeywords={Deep Learning, Time Series, Classification, Explainable AI, Discrete Representations}
}

\title{EXCODER: EXplainable Classification Of DiscretE time series Representations}
\renewcommand{\shorttitle}{EXCODER}

\author{
\href{https://orcid.org/0000-0003-4046-2990}{\includegraphics[scale=0.06]{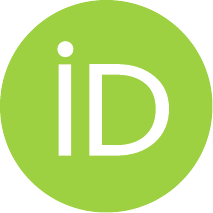}\hspace{1mm}Yannik Hahn}\thanks{These authors contributed equally.}\thanks{Corresponding author. {\it E-mail address:} yhahn@uni-wuppertal.de}, 
\href{https://orcid.org/0009-0009-2757-9570}{\includegraphics[scale=0.06]{orcid.pdf}\hspace{1mm}Antonin Königsfeld}\footnotemark[1], 
\href{https://orcid.org/0000-0003-0080-6285}{\includegraphics[scale=0.06]{orcid.pdf}\hspace{1mm}Hasan Tercan}, 
\href{https://orcid.org/0000-0002-1969-559X}{\includegraphics[scale=0.06]{orcid.pdf}\hspace{1mm}Tobias Meisen} \\
\textit{{yhahn, koenigsfeld, tercan, meisen}@uni-wuppertal.de} \\
Institute for Technologies and Management of Digital Transformation (TMDT)\\ 
University of Wuppertal \\
Rainer-Gruenter-Straße 21, 42119 Wuppertal, Germany \\ 
}

\date{Preprint, February 2026}
\renewcommand{\headeright}{A Preprint}
\renewcommand{\undertitle}{Accepted at PAKDD 2026}

\DeclareAcronym{VQ-VAE}{ short = VQ-VAE, long = vector quantised variational autoencoder }
\DeclareAcronym{DVAE}{ short = DVAE, long = discrete variational autoencoder }
\DeclareAcronym{VQ}{ short = VQ, long = vector quantisation }
\DeclareAcronym{VAE}{ short = VAE, long = variational autoencoder }
\DeclareAcronym{CNN}{ short = CNN, long = convolutional neural network }
\DeclareAcronym{DIN}{ short = DIN, long = German Institute for Standardisation Registered Association }
\DeclareAcronym{GMAW}{ short = GMAW, long = gas metal arc welding }
\DeclareAcronym{LSTM}{ short = LSTM, long = long short-term memory }
\DeclareAcronym{MAG}{ short = MAG, long = metal active gas }
\DeclareAcronym{MIG}{ short = MIG, long = metal inert gas }
\DeclareAcronym{RNN}{ short = RNN, long = recurrent neural network }
\DeclareAcronym{ml}{ short = ML, long = machine learning }
\DeclareAcronym{dl}{ short = DL, long = deep learning }
\DeclareAcronym{MLP}{ short = MLP, long = multilayer perceptron }
\DeclareAcronym{GRU}{ short = GRU, long = gated recurrent unit }
\DeclareAcronym{CV}{ short = CV, long = computer vision }
\DeclareAcronym{NLP}{ short = NLP, long = natural language processing }
\DeclareAcronym{IG}{ short = IG, long = Integrated Gradients }
\DeclareAcronym{SM}{ short = SM, long = Saliency Map }
\DeclareAcronym{SHAP}{ short = SHAP, long = SHapley Additive exPlanations }
\DeclareAcronym{LIME}{ short = LIME, long = Local Interpretable Model-agnostic Explanations }
\DeclareAcronym{Grad-SHAP}{ short = Grad-SHAP, long = Gradient SHAP }
\DeclareAcronym{DeepLIFT}{ short = DeepLIFT, long = Deep Learning Important FeaTures }
\DeclareAcronym{DeepSHAP}{ short = DeepSHAP, long = Deep SHAP }
\DeclareAcronym{RISE}{ short = RISE, long = Randomized Input Sampling for Explanation of Black-box Models }
\DeclareAcronym{LASTS}{ short = LASTS, long = Locally interpretable time series explanation }
\DeclareAcronym{CoMTE}{ short = CoMTE, long = Counterfactual Multivariate Time-series Explanations }
\DeclareAcronym{DELELSTM}{ short = DELELSTM, long = Decomposition-based Linear Explainable LSTM }
\DeclareAcronym{WinIT}{ short = WinIT, long = Window Importance for Time Series }
\DeclareAcronym{Grad-x-Input}{ short = Grad-x-Input, long = Gradient-times-Input }
\DeclareAcronym{AUC}{ short = AUC, long = Area Under the Curve }
\DeclareAcronym{ATF}{ short = ATF, long = attention flow }
\DeclareAcronym{ATM}{ short = ATM, long = attention map }
\DeclareAcronym{SSA}{ short = SSA, long = similar subsequence accuracy }
\begin{document}
\maketitle
\begin{abstract}
Deep learning has significantly improved time series classification, yet the lack of explainability in these models remains a major challenge. While Explainable AI (XAI) techniques aim to make model decisions more transparent, their effectiveness is often hindered by the high dimensionality and noise present in raw time series data. In this work, we investigate whether transforming time series into discrete latent representations—using methods such as Vector Quantized Variational Autoencoders (VQ-VAE) and Discrete Variational Autoencoders (DVAE)—not only preserves but enhances explainability by reducing redundancy and focusing on the most informative patterns. We show that applying XAI methods to these compressed representations leads to concise and structured explanations that maintain faithfulness without sacrificing classification performance. Additionally, we propose Similar Subsequence Accuracy (SSA), a novel metric that quantitatively assesses the alignment between XAI-identified salient subsequences and the label distribution in the training data. SSA provides a systematic way to validate whether the features highlighted by XAI methods are truly representative of the learned classification patterns. Our findings demonstrate that discrete latent representations not only retain the essential characteristics needed for classification but also offer a pathway to more compact, interpretable, and computationally efficient explanations in time series analysis.
\end{abstract}

\section{Introduction}
Time series analysis is fundamental to numerous domains, including healthcare, finance, and industrial monitoring, where the ability to track patient vitals, forecast stock market trends, and predict equipment failures is critical. As the volume and complexity of temporal data continue to grow, accurate and interpretable time series classification becomes essential for deriving actionable insights and supporting informed decision-making. Recent advances in deep learning have led to state-of-the-art performance in time series classification by capturing complex, non-linear dependencies directly from raw data, often outperforming traditional approaches. However, despite their predictive power, these models remain largely opaque, making it difficult to understand how decisions are made. This lack of transparency poses a significant challenge, particularly in high-stakes applications where trust and interpretability are paramount. Hence, understanding the reasoning behind their predictions is crucial, not only for debugging and refinement but also for ensuring robustness and reliability.

XAI methods address this challenge by making model decisions more transparent, enabling practitioners to identify key features, diagnose errors, and improve model trustworthiness. However, applying XAI to time series data introduces unique challenges due to the high dimensionality, temporal dependencies, and noise inherent in raw time series. These factors have driven a growing need for XAI techniques specifically designed for time series analysis, capable of providing meaningful and structured explanations without sacrificing predictive performance. 

Despite the growing need for explainability in time series classification, most existing XAI methods applied in this domain are adaptations of techniques originally developed for other data modalities, such as images or tabular data. For instance, techniques like Integrated Gradients (IG) \cite{sundararajan2017axiomatic}, LIME \cite{lime2016}, and attention mechanisms \cite{vaswani2017attention} have been extended to identify influential time points or subsequences in time series data. In an effort to structure the landscape of XAI for time series, \citet{theissler2022explainable} proposed a taxonomy that categorizes methods based on the level of granularity in their explanations. They distinguish between point-based approaches, which assess the importance of individual time steps, and subsequence-based approaches, which identify meaningful temporal segments to capture local patterns and dependencies. However, both categories come with inherent limitations: point-based methods may fail to consider broader temporal context, while subsequence-based approaches might lack the fine-grained resolution needed to capture short but critical variations in the data.

Beyond the adaptation of existing XAI methods, a complementary approach to improve interpretability in time series classification lies in learning structured representations of time series data. Discrete latent representations, such as those learned by Vector-Quantized Variational Autoencoders (VQ-VAE)~\cite{NIPS2017_7a98af17} and Discrete Variational Autoencoders (DVAE)~\cite{NEURIPS2018_9f53d83e}, have been successfully applied to various tasks, including anomaly detection, forecasting, and data compression. These models map continuous time series into a lower-dimensional space using a finite set of discrete codes, leading to more structured and compact representations. \citet{NEURIPS2018_9f53d83e} highlight several advantages of this approach, including noise reduction, improved robustness, and particularly enhanced feature interpretability, as discrete representations can enforce a more meaningful factorization of latent features. While these methods were not originally designed with explainability in mind, their ability to generate structured representations suggests they could serve as a foundation for more interpretable deep learning models.

Building on these advantages, recent work by \citet{zhao2018unsupervised, shvo2021interpretable} suggests that explanations generated within a discrete latent space can be more structured and actionable, as they operate on a compressed feature space that filters out noise and emphasizes essential patterns. Hence, in this work, we explore the potential of discrete latent representations to enhance the explainability of time series classification models. Specifically, we propose a patching mechanism, inspired by \citet{Yuqietal-2023-PatchTST}, that maps each discrete latent code back to its corresponding subsequence in the original time series. This ensures a direct and interpretable connection between abstract latent representations and the original input data. By integrating this approach with established XAI techniques, we systematically investigate whether discrete representations can enhance the interpretability of deep learning models for time series classification.

However, evaluating explanations in a discrete latent space calls for a different approach than traditional XAI evaluation methods, which are typically designed for high-dimensional continuous input data. To address this, we introduce Similar Subsequence Accuracy (SSA), a novel metric designed to quantify how well an XAI method identifies meaningful patterns within the latent space. SSA measures whether the most salient subsequences, as identified by an XAI technique, correspond to meaningful and recurring patterns in the training data. By doing so, it provides a quantitative framework for assessing the consistency and relevance of explanations in discretized representations.

To evaluate this, we explore the application of XAI methods on discrete latent representations of time series and compare their effectiveness to traditional approaches that operate directly on the original, continuous input space.

Summarized, the main contributions of this paper are:
\begin{enumerate}
    \item We adapt and implement a suite of XAI methods to operate effectively on discrete latent representations of time series, demonstrating their effectiveness in generating meaningful and interpretable explanations within this transformed space.
    \item We show that XAI methods applied to discrete representations not only maintain  performance comparable to those applied to the original input space but also produce significantly more compact explanations, thereby enhancing human interpretability and reducing explanation complexity.
    \item We introduce Similar Subsequence Accuracy (SSA), a novel metric for comparing XAI results across latent and original input spaces. SSA quantitatively assesses whether the most salient subsequences identified by an XAI method align with meaningful patterns in the training data, enabling a structured evaluation of explanation consistency and relevance.
\end{enumerate}

The remainder of this paper is structured as follows: We first discuss related work, positioning our approach within the landscape of existing methods for explainable time series classification. Next, we detail our methodology, covering data preprocessing, the adaptation of XAI methods to discrete representations, and the implementation of our proposed SSA metric. We then present the experimental setup and results, analyzing the key findings and their implications. Finally, we conclude by discussing the broader impact of our findings, identifying limitations, and suggesting directions for future research.
\section{Related Work}\label{sec:related_work}
\subsection{XAI Methods}\label{subsec:related_work_xai_methods}
Existing XAI methods for time series classification can be broadly categorized into gradient-based, perturbation-based, and attention-based approaches, each providing distinct insights into model behavior and feature importance.

Gradient-based methods exploit the gradients of a model's output with respect to the input to determine feature importance. Prominent approaches include Gradient explanations~\cite{baehrens2010explain} or \ac{SM}~\cite{DBLP:journals/corr/SimonyanVZ13}, \ac{IG}~\cite{sundararajan2017axiomatic}, GradientSHAP~\cite{erion2019learning}, DeepLIFT~\cite{shrikumar2017learning}, and DeepSHAP~\cite{lundberg2017unified}. Additionally, Gradient$\times$Input~\cite{Ancona2017TowardsBU} computes importance scores by multiplying input features with their corresponding gradients, highlighting the most influential regions within the time series. However, gradient-based methods can struggle to capture long-range temporal dependencies and complex non-linear relationships within time series data, which may limit their effectiveness for time series classification tasks where these characteristics are prevalent.

Modern deep learning architectures, particularly transformer-based models, inherently utilize attention mechanisms, which can be leveraged to provide explainability. While \citet{jain2019attention} questioned whether attention weights directly correspond to model decisions, \citet{wiegreffe2019attention} argued that they remain a valuable interpretability tool when properly contextualized. In this work, we refer to these techniques as Attention Maps (ATM). Furthermore, \citet{abnar2020quantifying} proposed methods, such as attention rollout and \ac{ATF}, to approximate the attention given to input tokens. In the context of time series classification, LASTS~\cite{guidotti_explaining_2020} generates explanations by identifying shapelets and providing exemplars and counter-exemplars that illustrate the reasoning behind the classification.

Post-hoc model-independent methods provide an alternative approach to explainability by generating local explanations without requiring modifications to the model itself. Techniques such as LIME\cite{lime2016} and RISE\cite{rise2018} achieve this by perturbing input features and analyzing the corresponding changes in the model’s predictions. These methods are particularly useful for gaining instance-specific insights but can struggle with the nature of time series data, where perturbations may disrupt temporal dependencies and lead to unreliable explanations.

Counterfactual explanations generate hypothetical samples, similar to the original but of a different class, illustrating minimal changes that would alter a model's prediction. \citet{mothilal2020dice} proposed DiCE for generating diverse counterfactuals, balancing proximity and actionable diversity via an optimization formulation with feasibility constraints.  However, DiCE's design for tabular data may yield infeasible time series counterfactuals by independently modifying time points.

While widely used, applying these approaches to time series is challenging due to high dimensionality and temporal dependencies, potentially leading to fragmented or overly localized attributions that miss broader sequential patterns. This motivates exploring discrete latent representations as a structured alternative for enhanced explainability.

\subsection{XAI Methods Designed for Time Series}\label{subsec:related_work_xai_methods_ts}
While general XAI methods have been adapted for time series, as discussed in Section~\ref{subsec:related_work_xai_methods}, methods designed specifically for the temporal domain utilize the temporal nature by capturing time-dependent features. 

\citet{ismail2020benchmarking} introduced Temporal Saliency Rescaling (TSR), a two-step approach that assigns importance scores based on both time relevance and feature relevance. However, TSR aggregates scores at individual time points, which may obscure dependencies within subsequences. Similarly, LAXCAT~\cite{hsieh2021explainable} identifies critical variables and time intervals, while \citet{leung_temporal_2023} propose a decomposition of feature importance across time and variable dimensions. Despite these advancements, these methods predominantly operate at the level of individual data points or aggregated windows, limiting their ability to capture structured, sequential patterns. In contrast, our importance scores are computed on learned discrete representations, each inherently linked to a specific subsequence. This allows for more precise explanations that directly reflect the learned sequential nature of the time series data.

\citet{crabbe2021explaining} introduced a perturbation-based method that learns a mask to highlight relevant parts of a time series, addressing challenges related to high dimensionality and noise. Although effective, this method requires specialized perturbation techniques to preserve time-dependent patterns. In contrast, our method operates directly in the latent space, where temporal dependencies are inherently preserved through the encoding process of discrete representations. This enables the application of standard XAI techniques without requiring task-specific perturbation strategies, thereby simplifying the explainability process while maintaining structured, sequence-aware explanations.

\subsection{Evaluation of XAI Methods}
Various approaches exist for evaluating XAI methods and comparing their explanations. \citet{adadi2018peeking} discuss evaluation as part of a broader review of XAI techniques, while \citet{chromik2020taxonomy} focus on user studies as an evaluation framework Other works examine evaluation in specific domains \cite{markus2021role}, local explanations \cite{guidotti2021evaluating} or counterfactual explanations \cite{stepin2021survey}. However, as \citet{doshi2018considerations} emphasize, user studies often favor explanations that pass an initial 'face-validity' test—appearing intuitive to humans—without necessarily ensuring their correctness or generalizability. Similarly, anecdotal evaluations risk cherry-picking examples that seem plausible but lack rigorous validation, limiting their reliability for assessing explanation quality.

To ensure a more systematic and objective evaluation, we adopt quantitative evaluation methods as proposed by \citet{zhou2021evaluating} and especially \citet{nauta2023anecdotal}. The latter introduced the "Co-12" properties, a framework for assessing explanation quality along multiple dimensions. These properties include:
\begin{enumerate}
\item Correctness, which measures how faithfully an explanation reflects the model’s reasoning, assessed via randomization checks for model parameters~\cite{adebayo2020debugging} and explanations~\cite{mohankumar2020towards}, as well as perturbation methods (e.g., single or incremental deletion)~\cite{bhatt2020evaluating}.
\item Output-completeness, evaluated using Preservation and Deletion Checks~\cite{dhurandhar2018explanations}.
\item Continuity, determined by testing explanation stability under slight variations in input data~\cite{alvarez2018towards}.
\item Coherence, assessed through agreement between different XAI methods~\cite{ancona2019explaining}.
\end{enumerate}

In this paper, we apply perturbation analysis (incremental deletion), implementation variance, and XAI method agreement. These evaluation methods are adaptable to discrete latent representations and cover a broad spectrum of available Co-12 properties, such as correctness, completeness, consistency, continuity, and coherence of explanations. We detail the evaluation techniques further in Section~\ref{subsec:evaluation_adaptations}.

\section{Methodology}
\subsection{Discrete Representations for Time Series Classification}
As previously discussed, we leverage discrete latent representations to enhance interpretability in time series classification. To achieve this, we focus on \ac{VQ-VAE} \cite{NIPS2017_7a98af17} as well as \ac{DVAE} \cite{NEURIPS2018_9f53d83e} and added SAX~\cite{lin2007experiencing} as a baseline. SAX divides the time series into equal-sized segments (PAA - Piecewise Aggregate Approximation), then maps these segments to discrete symbols using breakpoints that produce equiprobable regions under a Gaussian curve, creating a lower-bounding distance. \ac{VQ-VAE} and \ac{DVAE} learn a structured, discrete representation of time series data but differ in their quantization strategy:  \ac{VQ-VAE} maps continuous latent vectors to a learned codebook of discrete codes, while \ac{DVAE} models a categorical distribution over latent codes, enabling sampling via the Gumbel-Softmax trick ~\cite{jang2016categorical}. In both approaches, a raw time series \( x \) is transformed into a sequence of discrete tokens \( z \in \{0, \ldots, K-1\}^T \), where \(K\) is the codebook size and \(T\) the sequence length.

A central aspect of both VQ-VAE and DVAE is their reliance on a reconstruction objective, where the models learn to reconstruct the original time series from their compressed, discrete latent codes, thereby minimizing the information loss from this quantization step. To retain temporal structure, we incorporate a patching mechanism, inspired by \citet{Yuqietal-2023-PatchTST}. This technique associates each latent code, z, with a 25-timestep segment of the original time series. The segment length of 25 timesteps is a carefully chosen compromise, balancing the need to capture fine-grained local features with the requirement for descriptive, explainable temporal features. Consequently, each discrete code captures a specific, localized temporal pattern.

In addition to reconstruction, the learned discrete representations are directly used for classification. Instead of processing raw time series data, the classifier operates on the sequence of discrete latent codes, providing a structured representation that may facilitate interpretability by reducing complexity and highlighting key temporal patterns.

\subsection{Adaptation of XAI Methods}
As previously discussed, existing XAI methods are primarily designed for continuous input data and require adaptations to remain effective in the discrete latent space. In the following, we outline how established XAI techniques are adapted to ensure meaningful explanations within this transformed representation. To analyze discrete latent representations obtained from \ac{VQ-VAE} and \ac{DVAE}, we adapt three commonly used XAI techniques: \ac{SM}, \ac{IG}, and \ac{RISE}. A key challenge in this adaptation is the concept of masking, which plays a crucial role in IG and RISE but must be redefined for discrete token-based representations.

Since each token in the latent space corresponds to specific features of the original time series, the standard approach of using an uninformative baseline—such as zero-padding in image data—is not directly applicable. Instead, we introduce an \textit{unknown} token, analogous to the \textit{MASK} token in BERT~\cite{kenton2019bert}, to serve as a neutral placeholder during masking. This ensures minimal disruption to the learned representations while still allowing us to probe model behavior.

To seamlessly integrate the \textit{unknown} token, we modify the training process. Since both \ac{VQ-VAE} and \ac{DVAE} embed discrete tokens into a continuous latent space, the \textit{unknown} token is incorporated as an additional embedding, ensuring compatibility within the model’s input space.

\textbf{\ac{SM}:} For \ac{SM}, we adapted the method to operate on the discrete tokens by summing the gradients associated with each embedding vector. Let \( e_i \in \mathbb{R}^d \) be the embedding vector corresponding to the \( i \)-th token with a dimension of \(d\). The model's output for a given class c is denoted as \( S_c(z) \). The saliency value \( s_i \) for the \( i \)-th token is computed as follows:
\begin{equation}
s_i = \sum_{j=1}^{d} \left| \frac{\partial S_c(z)}{\partial e_{i,j}} \right| 
\end{equation}
where \( e_{i,j} \) is the \( j \)-th element of the embedding vector \( e_i \).

\textbf{\ac{IG}:} \ac{IG} requires a neutral baseline to accumulate gradients along a path from the baseline to the input token.  We introduce the \textit{unknown} token during training, replacing tokens with a probability of \( p_{unk} \in \{0.1, 0.15\} \), similar to noise-injection strategies in NLP models ~\cite{kenton2019bert}.

Let \( z' \) denote the baseline sequence where all tokens are replaced with the \textit{unknown} token, and \( e' \) its corresponding embedding representation. We compute the distances \( \delta(e_i, e') \) between token embedding \( e_i \) and \( e' \). These distances guide the path for integrating gradients. For each token, we define a path with \( m \) steps, \( \alpha = 1, \ldots, m \), to the token embedding \(e_i\) using linear interpolation. The integrated gradient \( I_i \) for the \( i \)-th token is then:
\begin{equation}
    IG_i = (e_i - e') \cdot \sum_{\alpha=1}^{m} \frac{\partial S_c(z^{(\alpha)})}{\partial z^{(\alpha)}} \times \frac{1}{m}
\end{equation}
\begin{equation}
    z^{(\alpha)} = e' + \frac{\alpha}{m}(e_i - e')
\end{equation}
where \( \frac{\partial S_c(z^{(\alpha)})}{\partial z^{(\alpha)}} \) represents the gradient of the model's output with respect to the interpolated embedding at step \( \alpha \)~\cite{sundararajan2017axiomatic}.

\textbf{\ac{RISE}:} For RISE, we extend the masking mechanism using the \textit{unknown} token to randomly perturb representations.  Each token is replaced with the \textit{unknown} token following a predefined probability distribution, and the impact on model predictions is measured. Let \( M \in \{0, 1\}^T \) be a binary mask where \( M_i = 0 \) indicates replacement of the \( i \)-th token with the \textit{unknown} token. The RISE importance score \( r_i \) for token \( i \) is:

\begin{equation}
 R = \frac{1}{|\mathcal{M}|}\sum_{M \in \mathcal{M}} P(c | z \odot M) \cdot M 
\end{equation}
where \( \mathcal{M} \) is a set of random masks, \( z \odot M \) represents the element-wise multiplication of the input sequence and the mask, and \(P(c | z \odot M) \) is the probability of the class c given the masked input ~\cite{rise2018}.

\textbf{\ac{LIME} and \ac{ATM}:} Unlike SM, IG and RISE, LIME and ATM require no modifications, as they naturally operate on discrete tokens.

\subsection{Evaluation Adaptations}
\label{subsec:evaluation_adaptations}
To ensure a fair comparison between models operating on latent and non-latent representations, we adapt the evaluation procedures accordingly. 

\subsubsection{Perturbation Analysis}
Perturbation analysis is used to validate whether the features identified by XAI methods are truly important for the model's decision-making process. By systematically altering the input and measuring the impact on predictions, we assess how strongly the model relies on the highlighted features. Hence, following \citet{nauta2023anecdotal}, we systematically perturb the discrete latent representations and measure the resulting impact on model performance.
\begin{enumerate}
    \item Non-latent models: Perturbation is applied to the raw time series by replacing a segment with a designated baseline value (e.g., zero values).
    \item Latent models: We replace a sequence of discrete tokens with the \textit{unknown} token, ensuring structural consistency within the representation.
\end{enumerate}

To establish a baseline for comparison, we introduce random perturbation (RND), where perturbations occur at randomly selected positions. We compute the \ac{AUC} score \( \text{AUC}_{RND} \) based on the F1 score across all classes, reflecting the model's robustness to arbitrary perturbations.

For targeted perturbations, we rank features by importance according to each XAI method, denoted as \( S = \{s_1, s_2, ..., s_n \} \), where \( s_1 \) is the most important feature.  We then incrementally perturb the highest-ranked features and compute the corresponding AUC score \( \text{AUC}_{XAI} \).

To quantify the relative effectiveness of XAI-guided perturbations, we compute:
\begin{equation}
\text{AUC}_{XAI-RND} = \text{AUC}_{RND} - \text{AUC}_{XAI}
\end{equation}

Thereby, a higher \( \text{AUC}_{XAI-RND} \) indicates that the XAI method successfully identifies features with a strong influence on model decisions. If an XAI method correctly identifies important features, we expect a significant performance degradation when those features are perturbed.

\subsubsection{Implementation Invariance}
\label{subsubsec:implementation_invariance}
To evaluate the stability of explanations, we test for implementation invariance following \citet{nauta2023anecdotal}. This ensures that feature attributions remain consistent across different training runs of functionally equivalent models. We train each model five times with five different randomly selected seeds, introducing variations in weight initialization and training dynamics while keeping the architecture unchanged. To quantify the stability of explanations, we compute the Cosine Similarity between feature importance scores generated by each XAI method. A high mean Cosine Similarity indicates that the explanations remain invariant to changes in initialization, supporting implementation invariance for the given model and explanation method.

\subsubsection{XAI Methods' Agreement}
To measure the consistency of explanations across different XAI methods, we compare feature importance rankings between methods applied to the same model. A high agreement suggests that multiple methods highlight similar features, increasing confidence in the reliability of explanations. We therefore compute the Cosine Similarity (CS) between feature importance scores obtained from multiple XAI methods. By comparing the explanations generated by methods such as \ac{SM}, \ac{LIME} and \ac{RISE}, we assess if they provide consistent interpretations. While a high agreement can indicate strong feature importance signals, disagreement between methods is not necessarily a negative outcome. Some XAI methods capture different perspectives on model behavior, which may be valuable in specific applications. However, if explainability relies too heavily on the choice of the XAI method rather than the model’s actual decision-making process, this may indicate instability in feature attribution. Ultimately, assessing XAI method agreement provides insight into the robustness of explanations, helping determine whether model interpretations are stable across different techniques. This is particularly relevant for model selection, as a high agreement across XAI methods suggests that the highlighted features are truly influential rather than artifacts of a specific interpretability technique.

\subsubsection{Similar Subsequence Accuracy (SSA)}
\label{Similar_Subsequence_Accuracy}
While the previously introduced evaluation methods assess different aspects of explainability, they do not explicitly validate whether the identified features correspond to characteristic, class-discriminative subsequences in the training data. The use of discrete latent representations now enables a direct comparison between model explanations and the structured patterns learned from the data. To leverage this, we introduce \ac{SSA}, a metric designed to evaluate the alignment between XAI explanations and recurring patterns in the training data. \ac{SSA} quantifies how often the most influential subsequences identified by an XAI method reappear in correctly classified training instances of the same class. This enables a structured assessment of whether model explanations capture genuine decision-relevant patterns or are influenced by spurious correlations. To compute \ac{SSA} for a given XAI method, we follow the steps in algorithm \ref{ssa_algorithm}.

\begin{algorithm}
\caption{SSA Algorithm}\label{ssa_algorithm}
\begin{algorithmic}[1]
\Require  Dataset \(D_{latent}^{train}\) and \(D_{latent}^{test}\), trained model \(M\), XAI method \(XAI\), subsequence length \(l\), set of Classes \(C\)
\For{\(c \in C\)} // iterate all classes
    \State  \(Matches_c \leftarrow 0\)
    \State  \(Neighborhoods_c \leftarrow 0\)
    \State \(Z_c \leftarrow \) all instances of \(D_{latent}^{test}\) that belong to class c.

    \For{\((z, c) \in Z_c\)} 
        \State \(i \leftarrow \text{argmax}(XAI(M, z))\) // argmax of importance
        
        \State \(N(z) \leftarrow \emptyset\) // initialize neighborhood
        \For{\((z', y') \in D_{latent}^{train}\)}
            \If{\(z_{i:i+l} \in z'\)}  // check matching sequence
                \State \(N(z) \leftarrow N(z) \cup \{z'\}\)
                \If{\(y' =  c\)} // check ground truth label
                    \State \(N(z)_{match} \leftarrow N(z)_{match} \cup \{z'\}\)
                \EndIf
            \EndIf
        \EndFor

         \State  \(Matches_c \mathrel{{+}{=}} |N(z)_{match}|\)
         \State  \(Neighborhoods_c \mathrel{{+}{=}} |N(z)|\)
    \EndFor
    \State \(\text{SSA}_{\text{class}}(c) \leftarrow \frac{Matches_c}{Neighborhoods_c}\)
\EndFor
\State \Return \(\text{SSA}= \frac{1}{|C|}\cdot\sum_{i=1}^{|C|} \text{SSA}_{\text{class}}(c_i)\)
\end{algorithmic}
\end{algorithm}

A high SSA indicates that the subsequence identified as important by the XAI method is frequently present in time series of the same class within the training data, and rarely present in time series of other classes. This suggests that the subsequence is a strong indicator of the class and that the XAI method has successfully identified a meaningful pattern in the latent space. Conversely, a low SSA suggests that the identified subsequence is not a reliable indicator of the class, either because it is not prevalent in the training data or because it is also common in other classes. Similarly to the perturbation analysis, we additionally compute the score on a random baseline highlighting arbitrary subsequences. Due to the heavy preprocessing and temporal dependencies in our datasets, we only focus on neighbors that feature similar subsequences at the same positions.

In essence, SSA provides a quantitative measure of how well the discrete latent representations capture class-specific patterns that are consistent with the historical data. With regards to the Co-12 properties defined by \citet{nauta2023anecdotal}, SSA measures explanation correctness by taking the alignment between true labels and model predictions into account, as well as consistency by ensuring that similar subsequences in the latent space lead to consistent predictions and explanations.
\section{Experiments}
\subsection{Datasets} \label{sec:datasets}
We use three time series classification datasets from industrial and medical domains to evaluate our approach. These datasets provide a diverse set of challenges, covering both univariate and multivariate time series and classification tasks with varying class imbalances. Each dataset contains at least \num{10000} training samples, ensuring robust model training and evaluation.

The Welding dataset~\cite{hahn_2025_15101072} provides multivariate time series from arc welding processes, focusing on quality prediction. It contains synchronously sampled current and voltage signals at 100 kHz, labeled as either "substandard" (43\%) or "satisfactory" (57\%) welds. 

The CNC dataset \cite{TNANI2022131} presents multivariate data from CNC machine operations. Tri-axial accelerometer data, sampled at 2 kHz, captures machine vibrations during normal and anomalous periods across various processes. For this study, we used a preprocessed version of the dataset, where a moving average filter was applied to reduce its size, followed by segmentation into smaller chunks.

The ECG dataset is a combination of the MIT-BIH Arrhythmia \cite{moody2001impact} and PTB Diagnostic ECG \cite{bousseljot1995nutzung} databases, preprocessed and segmented into individual heartbeats \cite{kachuee2018ecg}. This univariate dataset classifies heartbeats into five categories, including normal and various arrhythmia types.

We preprocess all datasets to ensure a consistent input format across models. Each time series is standardized to a uniform length, determined by the longest sequence within each dataset. This step ensures compatibility across latent and non-latent models. Table~\ref{tab:datasets} summarizes the dataset statistics.

\begin{table}[htb!]
\centering
\caption{Summary of dataset statistics where \#F denotes the number of features and \#C denotes the number of classes.}
\label{tab:datasets}
\begin{tabular}{l|cccc}
\toprule
\textbf{Dataset} & \textbf{Train/Val/Test} & \textbf{\#F} & \textbf{\#C}& \textbf{Class Distribution} \\
\midrule
Welding & 126k / 15k / 15k & 2 & 2 & \(57\% / 43\%\)\\
CNC & 13k / 1k / 1k & 3 & 2& \(4\% /~96\%\)\\
ECG & 87k / 10k / 10k & 1 & 5& \(83\%/ ~3\%/ ~7\% /~1\% /~7\%\)\\
\bottomrule
\end{tabular}
\end{table}

\begin{table}[htb!]
\centering
\caption{Combined results for all datasets. The table is structured with AUC metrics in the top rows and XAI methods as subcolumns within each model.  For each dataset, the highest values of \(AUC_{XAI-RND}\), \(AUC_{RND}\), and Cosine Similarity~(CS) Agreement are highlighted in bold.}
\resizebox{\textwidth}{!}{%
\begin{tabular}{l l |c c c c c | c | c }
\toprule
 &  & \multicolumn{5}{c|}{ \( \mathbf{AUC_{XAI-RND}} \) } & \( \mathbf{AUC} \) & \( \mathbf{CS} \) \\
Dataset & Model Name & SM & IG & RISE & LIME & ATM & RND & Agreement \\
\midrule
\multirow{8}{*}{CNC}& DLinear & \(0.0 \pm 0.0\) & \(0.0 \pm 0.0\) & \(-0.01 \pm 0.0\) & \(-0.0 \pm 0.0\) & - & \(0.49 \pm 0.0\) & \(0.84 \pm 0.0 \) \\
& DVAE MLP & \(0.03 \pm 0.03\) & \(-0.03 \pm 0.05\) & \(0.06 \pm 0.03\) & \(0.03 \pm 0.02\) & - & \(0.54 \pm 0.06\) & \(0.88 \pm 0.01 \)\\
& DVAE Tr & \(-0.01 \pm 0.01\) & \(0.0 \pm 0.0\) & \(0.0 \pm 0.0\) & \(-0.0 \pm 0.0\) & \(-0.01 \pm 0.02\) & \(0.5 \pm 0.01\) & \(0.78 \pm 0.01 \) \\
& MLP & \(-0.0 \pm 0.01\) & \(0.03 \pm 0.0\) & \(-0.0 \pm 0.01\) & \(0.01 \pm 0.01\) & - & \(\mathbf{0.65 \pm 0.0}\) & \(0.85 \pm 0.0 \) \\
& TS Tr & \(0.0 \pm 0.0\) & \(0.0 \pm 0.0\) & \(0.0 \pm 0.0\) & \(0.0 \pm 0.0\) & \(0.0 \pm 0.0\) & \(0.49 \pm 0.0\) & \(0.83 \pm 0.01 \) \\
& TimesNet & \(0.0 \pm 0.01\) & \(0.04 \pm 0.01\) & \(0.01 \pm 0.0\) & \(0.02 \pm 0.01\) & - & \(0.63 \pm 0.02\) & \(0.75 \pm 0.02 \) \\
& SAX MLP & \(0.01 \pm 0.01\) & \(0.01 \pm 0.01\) & \(0.01 \pm 0.01\) & \(0.04 \pm 0.01\) & - & \(0.54 \pm 0.01\) & \(\mathbf{0.89 \pm 0.01} \) \\
& VQ-VAE MLP & \(\mathbf{0.08 \pm 0.03}\) & \(-0.03 \pm 0.05\) & \(0.06 \pm 0.01\) & \(-0.02 \pm 0.02\) & - & \(0.55 \pm 0.07\) & \(0.8 \pm 0.01 \) \\
& VQ-VAE Tr & \(-0.05 \pm 0.01\) & \(0.01 \pm 0.01\) & \(0.0 \pm 0.01\) & \(-0.02 \pm 0.0\) & \(-0.01 \pm 0.05\) & \(0.55 \pm 0.01\) & \(0.78 \pm 0.02 \) \\
\midrule
\multirow{8}{*}{ECG}& DLinear & \(0.06 \pm 0.01\) & \(0.03 \pm 0.01\) & \(0.12 \pm 0.01\) & \(0.06 \pm 0.0\) & - & \(0.25 \pm 0.01\) & \(0.69 \pm 0.0 \) \\
& DVAE MLP & \(0.11 \pm 0.01\) & \(0.03 \pm 0.03\) & \(0.14 \pm 0.0\) & \(0.17 \pm 0.01\) & - & \(0.43 \pm 0.01\) & \(0.87 \pm 0.01 \) \\
& DVAE Tr & \(0.06 \pm 0.01\) & \(0.07 \pm 0.01\) & \(0.07 \pm 0.02\) & \(0.08 \pm 0.02\) & \(0.02 \pm 0.06\) & \(0.29 \pm 0.03\) & \(0.73 \pm 0.05 \) \\
& MLP & \(0.19 \pm 0.0\) & \(\mathbf{0.21 \pm 0.01}\) & \(0.2 \pm 0.01\) & \(0.12 \pm 0.0\) & - & \(0.43 \pm 0.02\) & \(0.71 \pm 0.02 \) \\
& TS Tr & \(0.09 \pm 0.02\) & \(0.12 \pm 0.02\) & \(0.04 \pm 0.03\) & \(0.01 \pm 0.03\) & \(0.0 \pm 0.12\) & \(0.33 \pm 0.01\) & \(0.7 \pm 0.06 \) \\
& TimesNet & \(0.07 \pm 0.01\) & \(0.09 \pm 0.01\) & \(0.11 \pm 0.0\) & \(0.01 \pm 0.01\) & - & \(0.39 \pm 0.01\) & \(0.69 \pm 0.0 \) \\
& SAX MLP & \(0.07 \pm 0.01\) & \(0.02 \pm 0.01\) & \(0.07 \pm 0.0\) & \(0.07 \pm 0.01\) & - & \(0.4 \pm 0.01\) & \(0.87 \pm 0.01 \) \\
& VQ-VAE MLP & \(0.15 \pm 0.05\) & \(0.04 \pm 0.07\) & \(0.12 \pm 0.02\) & \(0.17 \pm 0.02\) & - & \(\mathbf{0.44 \pm 0.03}\) & \(\mathbf{0.9 \pm 0.02}\) \\
& VQ-VAE Tr & \(0.11 \pm 0.05\) & \(0.08 \pm 0.01\) & \(0.14 \pm 0.02\) & \(0.11 \pm 0.02\) & \(0.06 \pm 0.04\) & \(0.36 \pm 0.02\) & \(0.76 \pm 0.01 \) \\
\midrule
\multirow{8}{*}{Welding}& DLinear & \(0.12 \pm 0.01\) & \(0.12 \pm 0.01\) & \(\mathbf{0.25 \pm 0.01}\) & \(0.02 \pm 0.0\) & - & \(0.52 \pm 0.0\) & \(0.69 \pm 0.0 \) \\
& DVAE MLP & \(0.07 \pm 0.03\) & \(0.01 \pm 0.01\) & \(0.1 \pm 0.02\) & \(0.12 \pm 0.03\) & - & \(0.64 \pm 0.03\) & \(0.87 \pm 0.01 \)\\
& DVAE Tr & \(0.04 \pm 0.04\) & \(0.03 \pm 0.01\) & \(0.11 \pm 0.01\) & \(0.16 \pm 0.02\) & \(0.07 \pm 0.04\) & \(\mathbf{0.65 \pm 0.02}\) & \(0.78 \pm 0.04 \) \\
& MLP & \(0.04 \pm 0.02\) & \(0.03 \pm 0.02\) & \(0.22 \pm 0.02\) & \(0.03 \pm 0.01\) & - & \(0.64 \pm 0.01\) & \(0.82 \pm 0.0 \) \\
& TS Tr & \(0.13 \pm 0.04\) & \(0.14 \pm 0.03\) & \(0.03 \pm 0.02\) & \(-0.03 \pm 0.06\) & \(0.07 \pm 0.05\) & \(0.57 \pm 0.05\) & \(0.65 \pm 0.09 \) \\
& TimesNet & \(0.05 \pm 0.02\) & \(0.03 \pm 0.01\) & \(0.21 \pm 0.01\) & \(-0.01 \pm 0.04\) & - & \(0.59 \pm 0.02\) & \(0.66 \pm 0.03 \) \\
& SAX MLP & \(0.02 \pm 0.02\) & \(0.0 \pm 0.02\) & \(0.1 \pm 0.01\) & \(0.1 \pm 0.03\) & - &  \(0.63 \pm 0.02\) & \(\mathbf{0.88 \pm 0.01} \)\\
& VQ-VAE MLP & \(0.09 \pm 0.06\) & \(0.05 \pm 0.03\) & \(0.14 \pm 0.01\) & \(0.09 \pm 0.02\) & - & \(0.6 \pm 0.03\) & \(0.87 \pm 0.01\) \\
& VQ-VAE Tr & \(0.02 \pm 0.03\) & \(0.03 \pm 0.01\) & \(0.11 \pm 0.02\) & \(0.09 \pm 0.01\) & \(0.05 \pm 0.04\) & \(0.63 \pm 0.03\) & \(0.76 \pm 0.01 \) \\
\bottomrule
\end{tabular}
}
\label{tab:combined_multirow_auc_xai_columns_results}
\end{table}

\subsection{Experimental Setup}
We evaluate the impact of XAI methods on both raw time series data and learned discrete latent representations. Our model selection encompasses a range of architectures, providing a comprehensive testbed across different representational spaces.

For the non-latent models, we evaluated four distinct architectures. DLinear~\citep{zeng2023transformers} and TimesNet~\citep{wu2022timesnet} were chosen as representatives of state-of-the-art time series models. An \ac{MLP} incorporating Layer Normalization and GeLU activation was included to represent a general-purpose neural network approach. Finally, a decoder-based time series Transformer was used to evaluate the impact of attention mechanisms.

We utilized \ac{VQ-VAE}~\citep{NIPS2017_7a98af17}, \ac{DVAE}~\citep{NEURIPS2018_9f53d83e}, and SAX~\citep{lin2007experiencing} to generate discrete latent representations. Classification on these latent spaces was performed using a Transformer architecture (following \citet{hahn2024quality}), chosen for its proven performance on similar data, and an \ac{MLP} for comparison.

The \ac{VQ-VAE} and \ac{DVAE} models were initially trained on a reconstruction task. Subsequently, all reconstruction and classification models underwent independent hyperparameter optimization. The optimal configuration for each classification model was then used to train five models with distinct random seeds. We report the mean and standard deviation of explainability metrics across these five runs to ensure robustness and reproducibility.

To achieve a broad evaluation, we incorporated a diverse set of five widely-used XAI methods, encompassing gradient-based, perturbation-based, and attention-based techniques. Specifically, we employed \ac{SM}, \ac{IG}, \ac{LIME}, \ac{RISE}, and \ac{ATM}. The best hyperparameters and the full code are available on GitHub\footnote{\url{https://github.com/tmdt-buw/EXCODER}}.

\subsection{Results} \label{sec:results_perturb}
The experiments aim to evaluate the robustness of latent and non-latent models under perturbations, analyze the agreement between different XAI methods, and assess the stability of explanations across multiple training runs.
Table~\ref{tab:combined_multirow_auc_xai_columns_results} presents the results of the perturbation analysis revealing that the effectiveness of a given XAI method varies depending on both the model architecture and the dataset. Notably, VQ-VAE combined with an \ac{MLP} using \ac{SM} achieves the highest score on the CNC dataset, while the \ac{MLP} with \ac{IG} outperforms other models on the ECG dataset, and DLinear with \ac{RISE} achieves the best performance on the Welding dataset. \ac{RISE} and \ac{LIME} appear to perform particularly well with latent models, achieving the best scores for this model type on the ECG and Welding datasets. These findings indicate that the efficacy of XAI methods is not uniform but instead depends on the interplay between the model and the data.

Column \textit{RND} of Table~\ref{tab:combined_multirow_auc_xai_columns_results} quantifies model robustness to input perturbations. This metric is influenced by factors such as the number of classes, sequence length, and data dimensionality. While non-latent models show higher robustness on the CNC dataset, latent models achieve comparable or better robustness on the ECG and Welding datasets. The absence of a consistently superior model across all datasets highlights the need to consider dataset-specific characteristics in model selection.

The last column of Table \ref{tab:combined_multirow_auc_xai_columns_results} illustrates the agreement between different XAI methods when applied to the same model and dataset. The latent models, particularly DVAE with an \ac{MLP} on the CNC and Welding datasets and VQ-VAE with an \ac{MLP} on the ECG dataset, exhibit the highest degree of agreement. This indicates a greater consistency in identified salient features by different XAI methods within the latent space compared to non-latent models. Specifically, the VQ-VAE \ac{MLP} model consistently shows high agreement among different XAI methods, indicating that the choice of XAI method might be less critical when using this particular model and that the model's predictions might be more robust to the specific XAI method employed.

\begin{figure}
    \centering
    \includegraphics[width=0.7\linewidth]{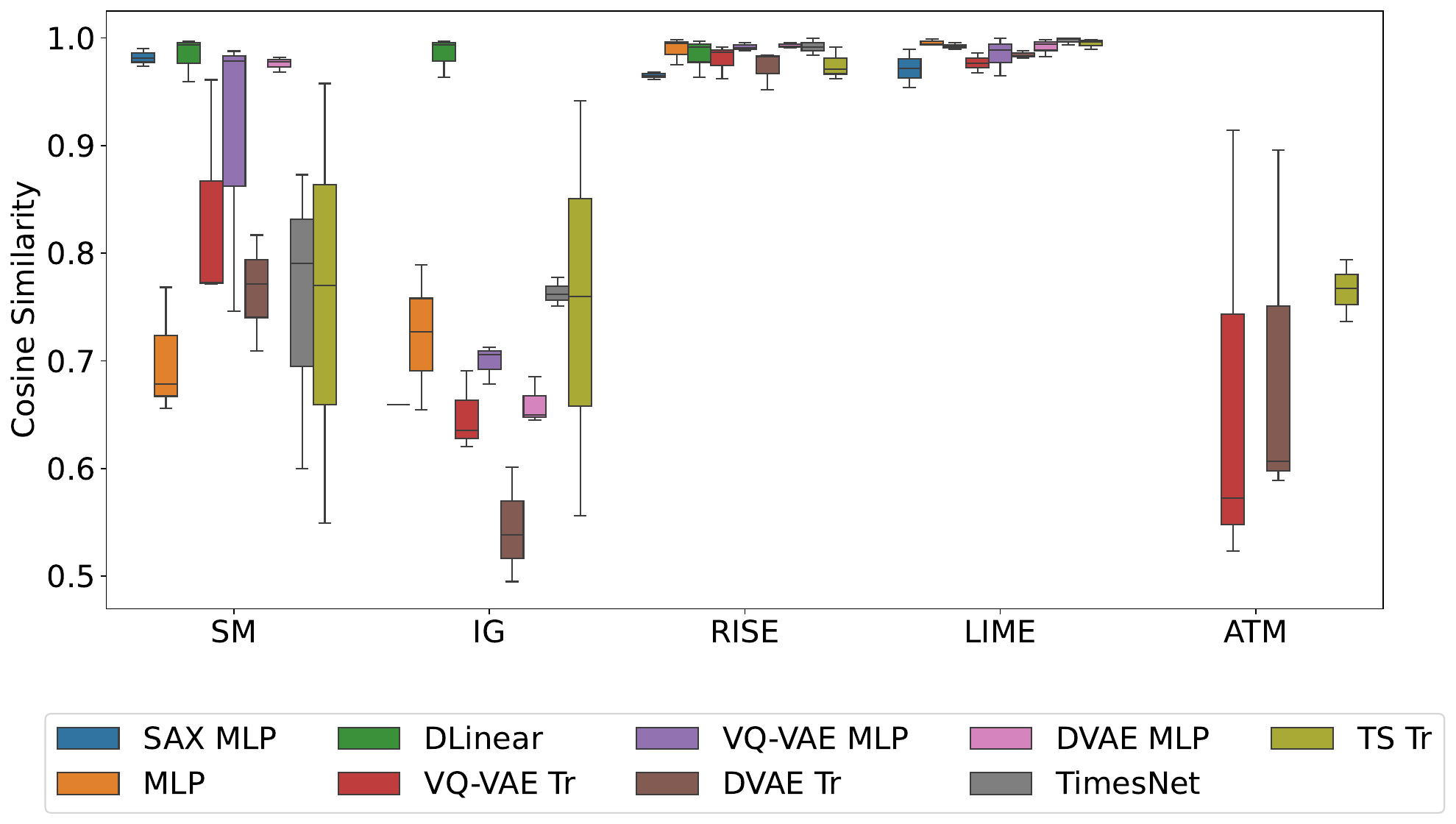}
    \caption{Implementation invariance of explanations over all datasets}
    \label{fig:boxplot_cosine_similarity}
\end{figure}

Figure \ref{fig:boxplot_cosine_similarity} displays the consistency of the XAI explanations across different random seeds, providing a measure of implementation invariance, as defined in section \ref{subsubsec:implementation_invariance}. While \ac{SM} and \ac{IG} exhibit higher variance across seeds, \ac{LIME} demonstrates the highest invariance, closely followed by \ac{RISE}. This indicates that post-hoc methods provide more consistent explanations than model-agnostic methods. 

\subsubsection{SSA Results}
\begin{figure}
    \centering
    \includegraphics[width=0.85\linewidth]{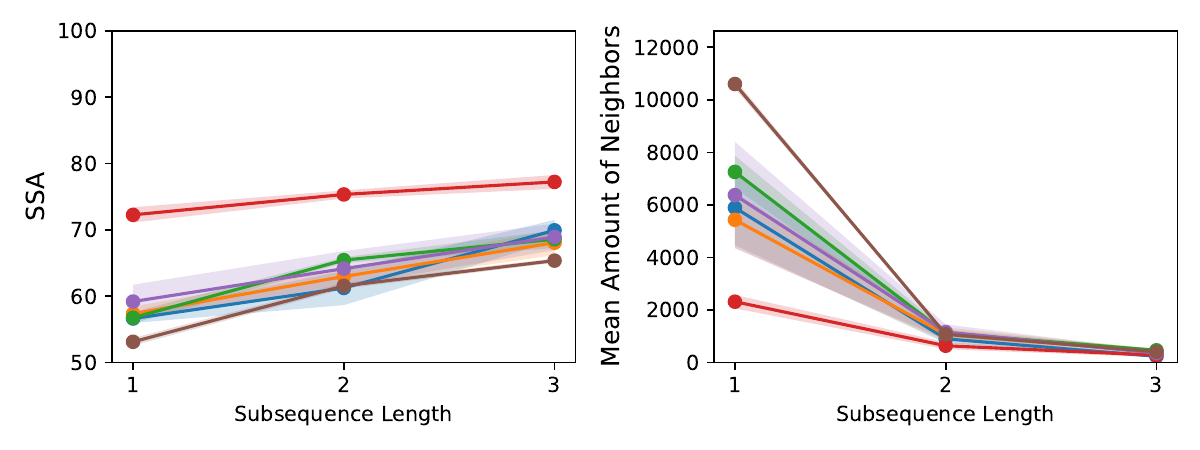}
    \caption{Exemplary SSA Scores with different subsequence lengths (VQ-VAE Transformer on Welding Dataset)}
\label{fig:ssa_scores_over_subsequence_lengths}
\end{figure}

In our evaluation, we consider subsequences of varying length. Figure~\ref{fig:ssa_scores_over_subsequence_lengths} illustrates how SSA scores evolve with increasing subsequence length, along with the mean number of neighbors per instance, as described in Section~\ref{Similar_Subsequence_Accuracy}. Naturally, the number of neighbors drastically decreases with a longer similar subsequence, therefore affecting the validity of SSA. This leads to the decision of including no subsequences longer than three tokens. Figure \ref{fig:ssa_scores_over_subsequence_lengths} implies that \ac{LIME} outperforms other XAI methods in capturing class-specific patterns when applied to the VQ-VAE Transfomer classifying the welding dataset. Furthermore, as expected, the random baseline exhibits the lowest SSA scores while producing larger instance neighborhoods. This confirms that high SSA values indicate more distinctive subsequences that differentiate between instances, aligning with the goal of high-quality explainability.

\begin{figure}
    \centering
    \includegraphics[width=0.9\linewidth]{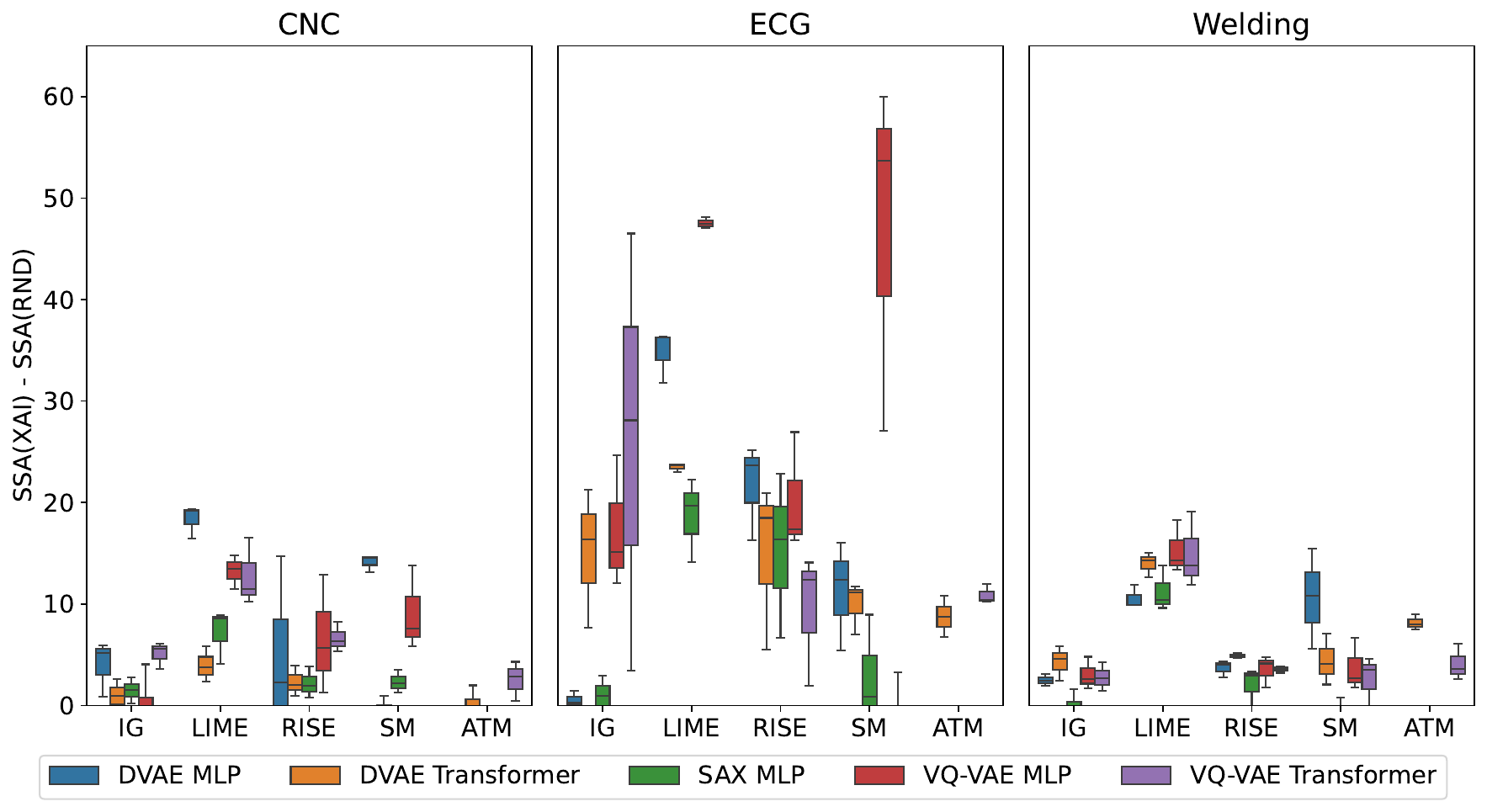}
    \caption{Distribution of SSA improvement over random baselines across the CNC, ECG, and Welding datasets. Boxplots show the difference between SSA achieved by each XAI method (SM, IG, LIME, RISE, ATM) and a random baseline (SSA(XAI) - SSA(RND)).}
    \label{fig:ssa_result_mean}
\end{figure}

In order to provide an overview for all tested model architectures and datasets, we compare a mean over three different subsequence lengths for each method to the random baseline. Figure \ref{fig:ssa_result_mean} indicates that according to the mean difference \(SSA(XAI)-SSA(RND)\), explanations provided by \ac{LIME} consistently outperform other XAI methods for all datasets and model architectures. Using \ac{LIME}, the VQ-VAE MLP has the best SSA score for the Welding and ECG dataset, while the DVAE MLP works best for the CNC dataset. 

\begin{figure}
    \centering
    \begin{subfigure}{0.32\linewidth}
        \includegraphics[width=\linewidth]{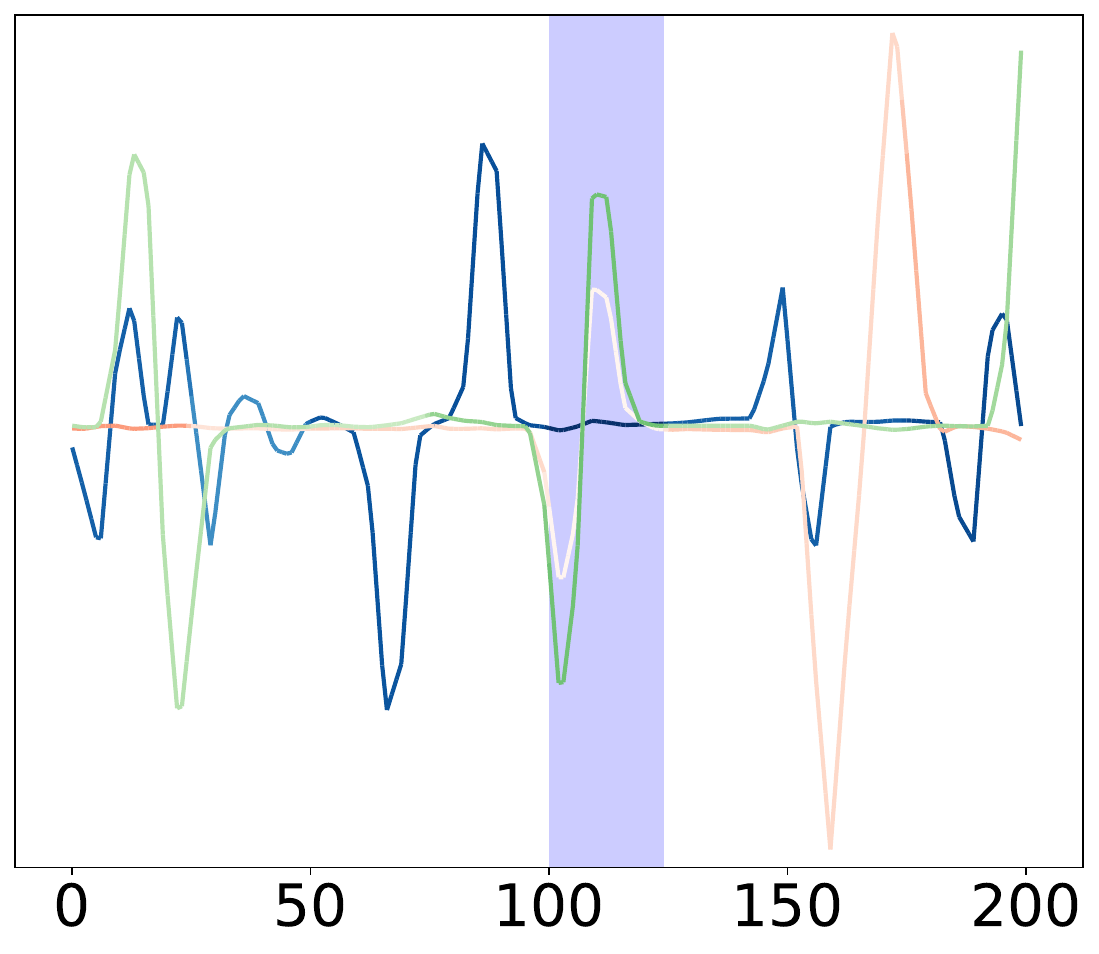}
        \caption{CNC Sample}
        \label{fig:CNC_Machining_VQ-VAE_MLP_SM}
    \end{subfigure}
    \hfill
    \begin{subfigure}{0.32\linewidth}
        \includegraphics[width=\linewidth]{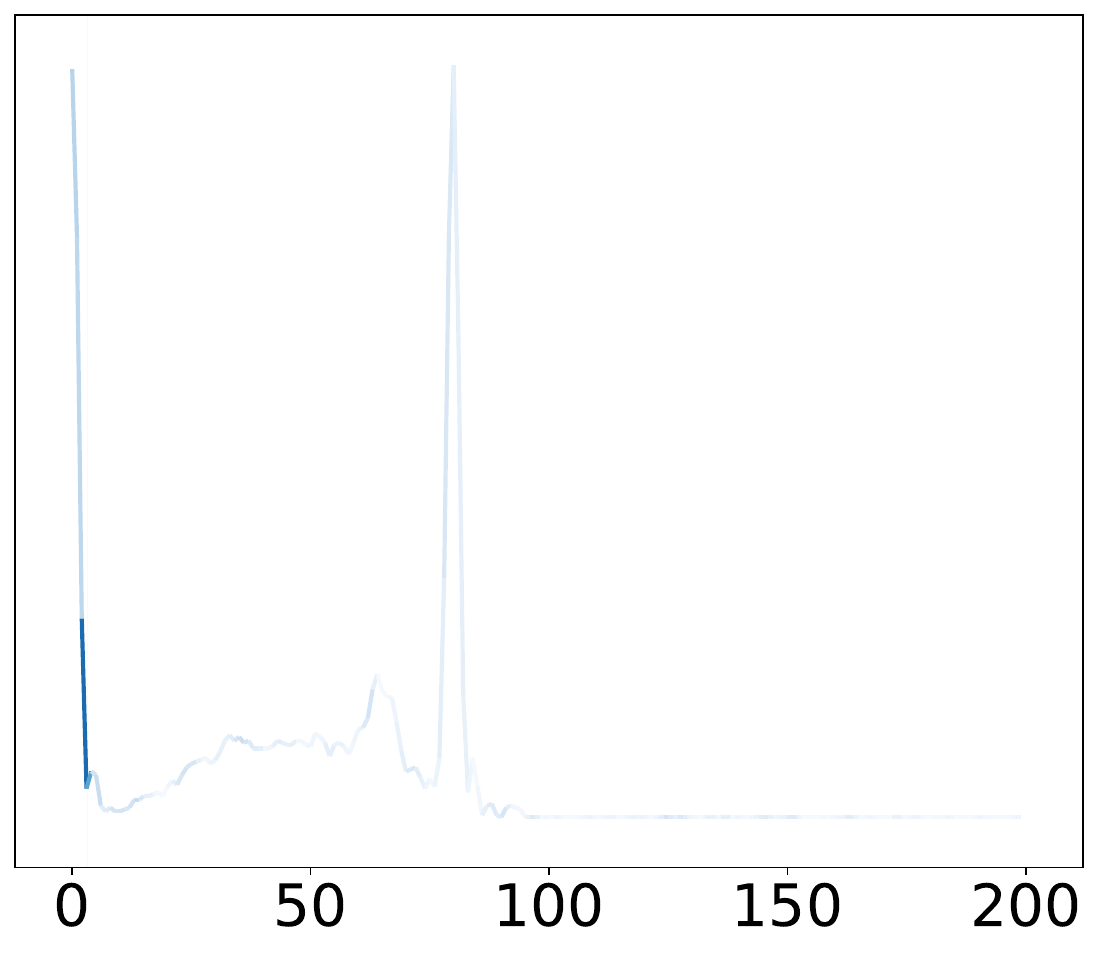}
        \caption{ECG Sample}
        \label{fig:ECG_MLP_IG}
    \end{subfigure}
    \hfill 
    \begin{subfigure}{0.32\linewidth}
        \includegraphics[width=\linewidth]{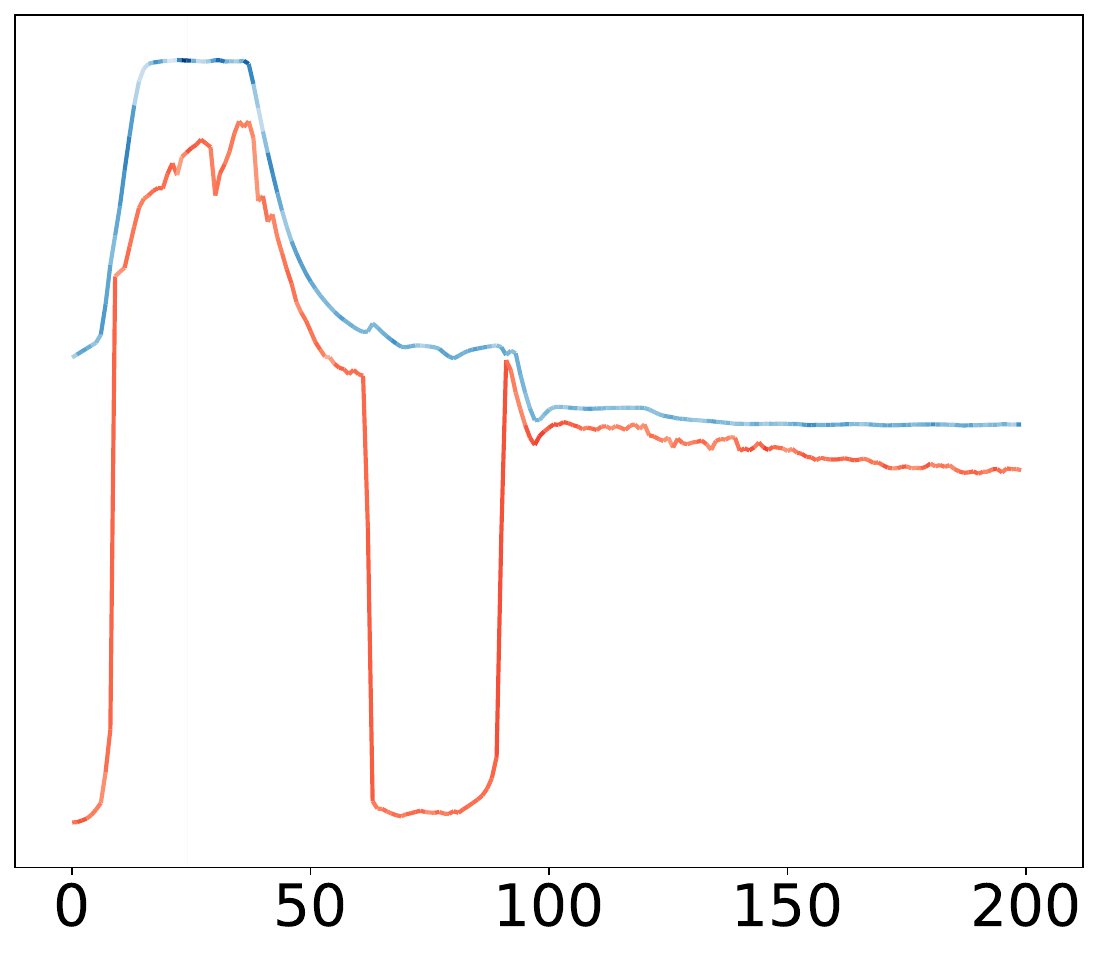}
        \caption{Welding Sample}
        \label{fig:Welding_DLinear_RISE}
    \end{subfigure}
    \vfill
    \begin{subfigure}{0.32\linewidth}
        \includegraphics[width=\linewidth]{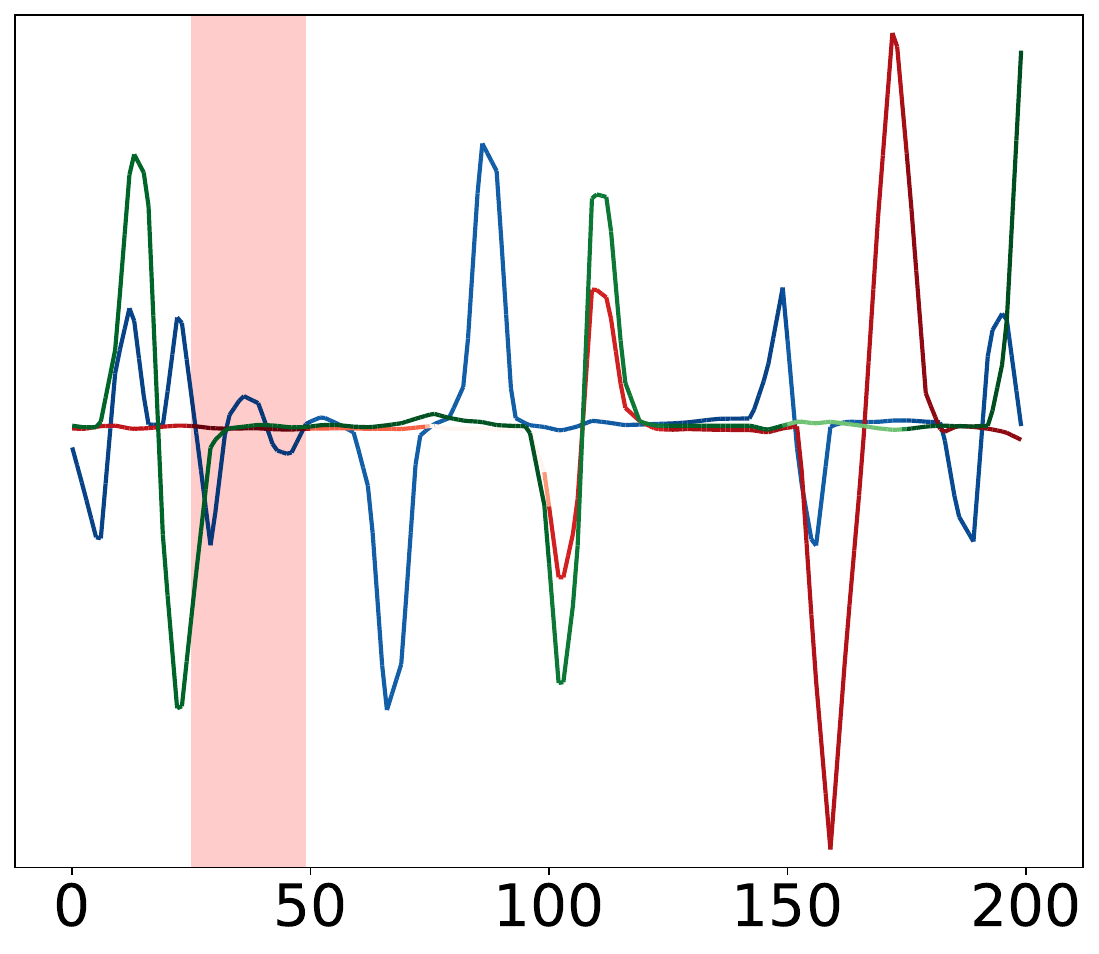}
        \caption{SSA: 93.82\% with 1941 neighbors} 
        \label{fig:CNC_Machining_DVAE_MLP_LIME}
    \end{subfigure}
    \hfill
    \begin{subfigure}{0.32\linewidth}
        \includegraphics[width=\linewidth]{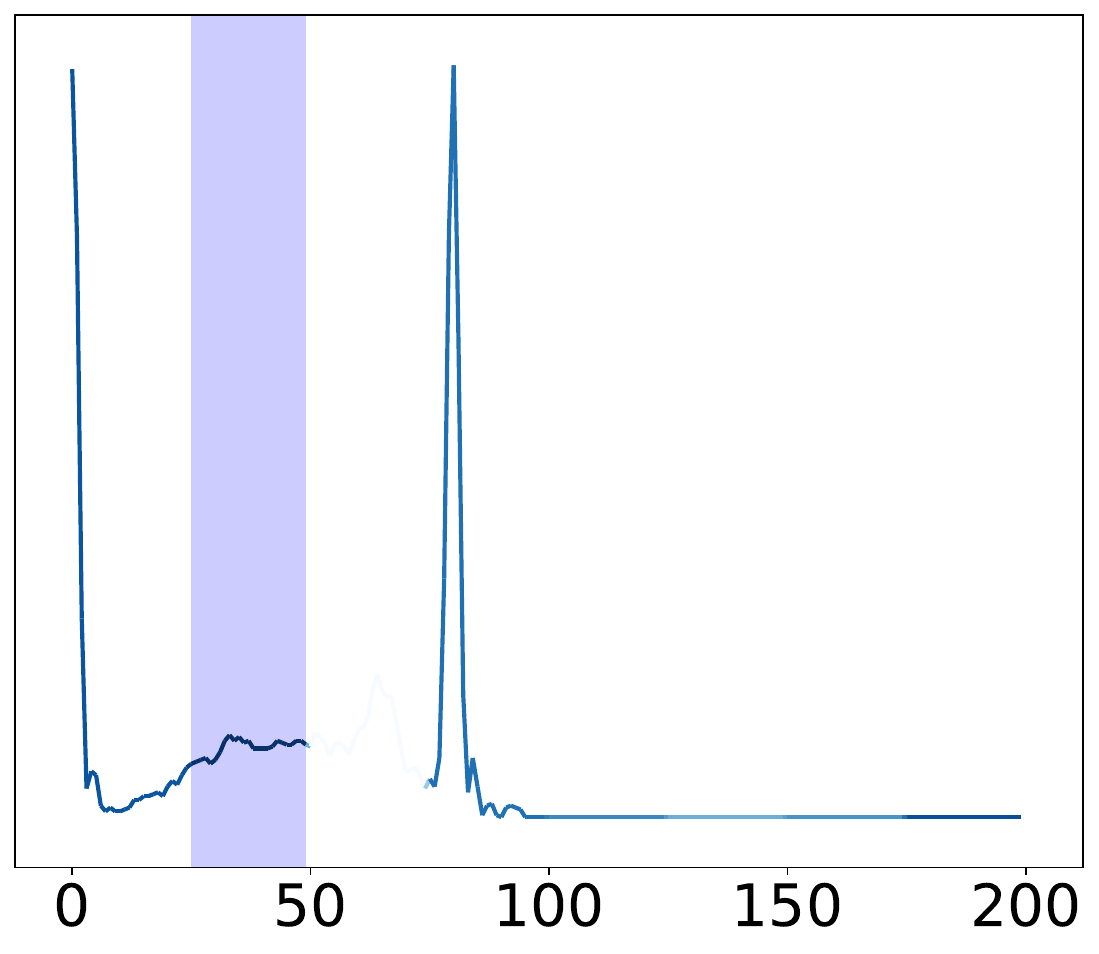}
        \caption{SSA: 98.9\% with 1464 neighbors}
        \label{fig:ECG_VQ-VAE_MLP_LIME}
    \end{subfigure}
    \hfill 
    \begin{subfigure}{0.32\linewidth}
        \includegraphics[width=\linewidth]{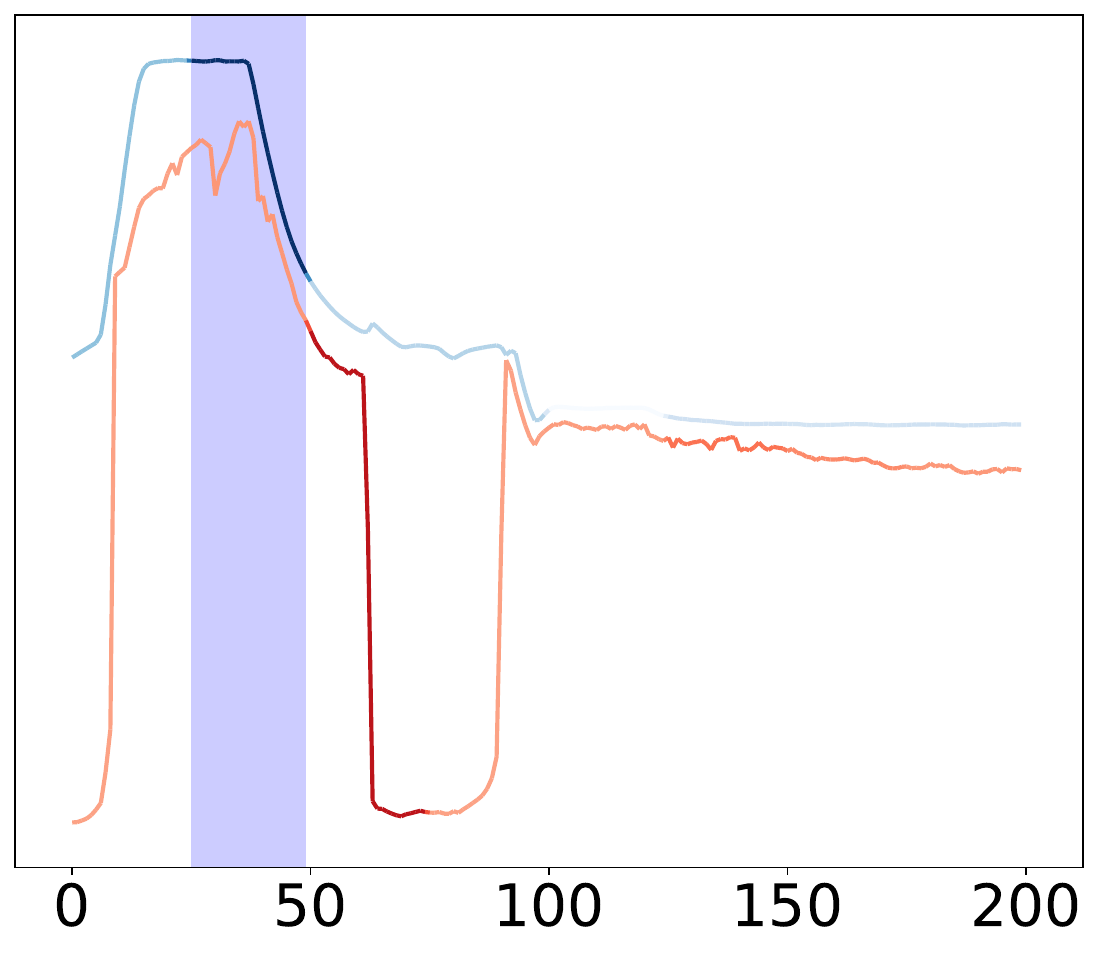}
        \caption{SSA: 83.3\% with 2322 neighbors}
        \label{fig:Welding_VQ-VAE_MLP_LIME}
    \end{subfigure}
    \caption{Explanations for random samples from CNC (a, d), ECG (b, e), and Welding (c, f) datasets. Top row (a-c): Explanations from best-performing models (via perturbation), with darker colors indicating higher feature importance. Bottom row (d-f): Explanations with highest Similar Subsequence Accuracy (SSA), highlighting the subsequence used for SSA calculation. SSA percentages and neighbor counts are shown below (d-f).}
    \label{fig:multiple_explanations}
\end{figure}

Figure~\ref{fig:multiple_explanations} demonstrates the advantages of subsequence-based explanations over point-wise approaches. While point-based methods (e.g., Figs.~\ref{fig:ECG_MLP_IG} and \ref{fig:Welding_DLinear_RISE}) often fail to clearly identify crucial features, the patch-based representations (e.g., Figs.~\ref{fig:CNC_Machining_VQ-VAE_MLP_SM}, \ref{fig:CNC_Machining_DVAE_MLP_LIME}, \ref{fig:ECG_VQ-VAE_MLP_LIME}, and \ref{fig:Welding_VQ-VAE_MLP_LIME}) highlight relevant subsequences with greater precision. This allows for a more accurate identification of important features within the time series. Moreover, the highlighted regions, combined with the SSA percentage (representing the prevalence of the same subsequence and label combination in the training data), provide a two-fold benefit: they indicate which part of the time series is important and offer insight into the reliability of the prediction based on the training data distribution.
\section{Conclusion and Outlook}
Recent advancements in deep learning for time series analysis have highlighted the critical need for explainability, particularly given the inherent black-box nature of many state-of-the-art models. While Explainable AI (XAI) provides methods to address this challenge, directly applying XAI to raw, continuous time series data poses several difficulties, including high dimensionality, noise, and the challenge of defining meaningful subsequences. To overcome these limitations, we proposed an approach that utilizes discrete latent representations to enhance the interpretability and effectiveness of XAI methods in time series classification.

This work makes several key contributions. First, we adapted established XAI methods—including gradient-based, perturbation-based, and attention-based techniques—to operate on discrete latent representations. Second, through extensive experimentation, we demonstrated that these adapted methods achieve comparable performance to traditional XAI approaches operating on raw time series, while producing significantly more compact and structured explanations. Third, we introduced SSA, a novel metric designed to assess whether the features identified by an XAI method correspond to class-specific patterns in the training data. This metric provides an objective measure of explanation consistency, ensuring that identified features genuinely represent meaningful class-discriminative information rather than misleading correlations.

The experimental results highlight several key findings. Firstly, there is no universally superior model-XAI combination, reinforcing the importance of dataset-specific evaluations. Secondly, a higher agreement between XAI methods in discrete latent spaces suggests that feature attributions in discrete representations are more stable. Thirdly, post-hoc perturbation-based methods (LIME, RISE) are less sensitive to model initialization than gradient-based methods (SM, IG). Finally, SSA shows that LIME has the best score for every dataset and models.

Despite these advancements, our work has certain limitations. The experiments were conducted on specific datasets and model architectures, and while the adapted XAI methods performed comparably to those applied to raw time series, no single method consistently outperformed others across all datasets. This suggests that XAI effectiveness remains model- and data-dependent, reinforcing the need for context-aware evaluation of explainability techniques.

Looking ahead, this work represents a first step towards integrating XAI with discrete time series representations. Future research should build on these findings by developing XAI techniques specifically designed for discrete latent spaces, which could lead to more refined and tailored interpretability methods. Additionally, a unified model that jointly learns the discrete representation and classification task could further streamline explainability by eliminating the need for separate encoding and classification steps. Finally, collaborations with domain experts will be essential to assess the practical utility of the generated explanations, ensuring their relevance and applicability in real-world scenarios.

\section*{GenAI Usage Disclosure}

Generative AI tools were used to support code development and assist in drafting parts of the text. All content was critically reviewed, edited, and remains the sole responsibility of the authors.

\bibliographystyle{plainnat}
\bibliography{ref}  

\end{document}